\documentclass[conference]{IEEEtran}
\IEEEoverridecommandlockouts

\usepackage{cite}
\usepackage{amsmath,amssymb,amsfonts}
\usepackage{algorithmic}
\usepackage{graphicx}
\usepackage{textcomp}
\usepackage{xcolor}
\def\BibTeX{{\rm B\kern-.05em{\sc i\kern-.025em b}\kern-.08em
    T\kern-.1667em\lower.7ex\hbox{E}\kern-.125emX}}

\usepackage{xcolor}
\usepackage{lipsum}
\usepackage{amsmath,amssymb}
\usepackage{array}
\usepackage{multirow}
\usepackage{booktabs}

\makeatletter
\newcommand{\linebreakand}{%
  \end{@IEEEauthorhalign}
  \hfill\mbox{}\par
  \mbox{}\hfill\begin{@IEEEauthorhalign}
}
\makeatother

\begin{document}

\title{Neuromorphic Cybersecurity with Semi-supervised Lifelong Learning
\thanks{This material is based upon work supported in part by the U.S. Department of Energy, Office of Science, Office of Advanced Scientific Computing Research, under Award Number \#DE-SC0021562.}
}


\author{
\linebreakand
    \IEEEauthorblockN{Md Zesun Ahmed Mia}
    \IEEEauthorblockA{\textit{School of EECS} \\
        \textit{The Pennsylvania State University}\\
        University Park, PA, USA  \\
        zesun.ahmed@psu.edu}
        \vspace{-8pt}
    \and
    \IEEEauthorblockN{Malyaban Bal}
    \IEEEauthorblockA{\textit{School of EECS} \\
        \textit{The Pennsylvania State University}\\
        University Park, PA, USA  \\
        mjb7906@psu.edu}
        \vspace{-8pt}
    \and
    \IEEEauthorblockN{Sen Lu}
    \IEEEauthorblockA{\textit{Dept. of EECS} \\
        \textit{University of Michigan}\\
        Ann Arbor, MI, USA \\
        senlu@umich.edu}
        \vspace{-8pt}
    \and
    \IEEEauthorblockN{George M. Nishibuchi} 
    \IEEEauthorblockA{\textit{Quantum Ventura, Inc.}\\ 
        San Jose, CA, USA \\ 
        max@quantumventura.com} 
        \vspace{-20pt}
\linebreakand
    \IEEEauthorblockN{Suhas Chelian} 
    \IEEEauthorblockA{\textit{University of Texas at Arlington Research Institute}\\ 
        Fort Worth, TX, USA \\ 
        suhas.chelian@uta.edu} 
        \vspace{-20pt}
    \and
    \IEEEauthorblockN{Srini Vasan} 
    \IEEEauthorblockA{\textit{Quantum Ventura, Inc.}\\ 
        San Jose, CA, USA \\ 
        srini@quantumventura.com} 
    \vspace{-20pt}
    \and
    \IEEEauthorblockN{Abhronil Sengupta}
    \IEEEauthorblockA{\textit{School of EECS} \\
        \textit{The Pennsylvania State University}\\
        University Park, PA, USA \\
        sengupta@psu.edu}
        \vspace{-20pt}
        }

\maketitle

\begin{abstract}
Inspired by the brain's hierarchical processing and energy efficiency, this paper presents a Spiking Neural Network (SNN) architecture for lifelong Network Intrusion Detection System (NIDS). The proposed system first employs an efficient static SNN to identify potential intrusions, which then activates an adaptive dynamic SNN responsible for classifying the specific attack type. Mimicking biological adaptation, the dynamic classifier utilizes Grow When Required (GWR)-inspired structural plasticity and a novel Adaptive Spike-Timing-Dependent Plasticity (Ad-STDP) learning rule. These bio-plausible mechanisms enable the network to learn new threats incrementally while preserving existing knowledge. Tested on the UNSW-NB15 benchmark in a continual learning setting, the architecture demonstrates robust adaptation, reduced catastrophic forgetting, and achieves $85.3$\% overall accuracy. Furthermore, simulations using the Intel Lava framework confirm high operational sparsity, highlighting the potential for low-power deployment on neuromorphic hardware.

\end{abstract}



\section{Introduction}
\IEEEPARstart{N}{etwork} Intrusion Detection Systems (NIDS) face significant challenges in scalability and energy efficiency, especially in high-throughput environments. The inherent nature of network attacks---often sparse events within a continuous temporal data stream---makes SNNs a compelling, bio-inspired alternative \cite{sengupta2019going} to tackle such an application driver. SNNs offer potential for energy savings and real-time processing due to their event-driven, sparse computation. Furthermore, real-world cybersecurity scenarios often lack comprehensive labeled data, necessitating unsupervised or semi-supervised learning approaches.
While Spike-Timing-Dependent Plasticity (STDP) is a common unsupervised learning rule in SNNs, its performance can be limited in complex classification tasks \cite{diehl2015unsupervised}. To enhance the efficacy and efficiency of STDP-based SNNs for NIDS, we explore a \textbf{hierarchical architecture} inspired from the brain's architectural organization \cite{ledoux2012rethinking}. This involves an initial, lightweight detection phase filtering benign traffic, followed by a more detailed classification phase, optimizing resource usage and potentially improving learning focus.
Beyond static threats, NIDS must contend with the dynamic nature of cyberattacks, requiring adaptation to novel threats over time---a challenge known as \textbf{lifelong learning}. Conventional networks, and often static SNNs, suffer from \textit{catastrophic forgetting}, losing old knowledge when learning new patterns \cite{kemker2018measuring}. Addressing this requires mechanisms for incremental learning.

To tackle both the STDP performance limitations and the lifelong learning challenge, this paper proposes and evaluates a \textbf{Hierarchical Dynamic Spiking Neural Network (D-SNN)}. This architecture combines the efficiency of the hierarchical structure with mechanisms for continuous adaptation. 
Learning employs our novel \textbf{Adaptive STDP (Ad-STDP)} rule, using a neuron's activity history (`firing factor') to modulate plasticity, stabilizing memories while learning new patterns \cite{diehl2015unsupervised}. We test this D-SNN on the UNSW-NB15 NIDS dataset, analyzing its adaptation, knowledge retention, efficiency, and neuromorphic hardware suitability via Lava simulations \cite{davies2021advancing}.



\section{Related Works and Main Contributions}

Neuromorphic approaches, particularly SNNs, are being explored for cybersecurity due to their potential for low-power, real-time processing. Prior work in neuromorphic NIDS includes systems demonstrating significant speed/energy improvements over conventional methods \cite{follett2017neuromorphic} and implementations on hardware like Intel's Loihi or using ANN-to-SNN conversion techniques \cite{zahm2022cyber}. Unsupervised methods using autoencoders on neuromorphic simulators or hardware have also shown promise \cite{alom2017network}. However, many existing neuromorphic NIDS often rely on supervised training paradigms or do not explicitly tackle the challenge of continuously adapting to new, unseen threats without forgetting past ones. Lifelong learning, or continual learning, aims to address this adaptation challenge, but mitigating catastrophic forgetting remains difficult, especially in SNNs \cite{ kemker2018measuring}. Strategies explored in SNNs include leveraging synaptic plasticity rules like STDP \cite{diehl2015unsupervised}, developing adaptive plasticity mechanisms \cite{panda2017asp}, controlling forgetting through neuromodulation or structural plasticity \cite{allred2020controlled}, and employing dynamic architectures inspired by concepts like GWR networks \cite{marsland2002self}. While these approaches show promise, integrating them effectively into a practical, efficient, and hierarchical NIDS framework remains an area ripe for investigation. Our work builds upon these foundations, combining a hierarchical SNN structure (justified in Sec.~I) with dynamic adaptation and adaptive learning rules---specifically, our novel Ad-STDP incorporating a \textbf{neuron-specific firing factor to modulate plasticity for lifelong learning, differing from prior adaptive mechanisms}---for robust lifelong NIDS, explicitly tackling limitations of prior works, particularly their tendency to overlook hierarchical scalability or rigorous lifelong learning evaluation in this context.
The main contributions of this paper are:
\begin{itemize}
    \item \textbf{Hierarchical SNN with Dynamic Lifelong Learning Classifier:} Design and application of a novel, bio-plausible hierarchical SNN architecture for NIDS. This features an efficient static SNN detector (using standard STDP) followed by an adaptive dynamic SNN classifier. The dynamic classifier employs GWR-inspired structural plasticity combined with Adaptive STDP (Ad-STDP) learning to enable continuous learning of new attack types while mitigating catastrophic forgetting.
    \item \textbf{Semi-Supervised Lifelong Evaluation:} Demonstration of the semi-supervised D-SNN's effectiveness on the UNSW-NB15 benchmark \cite{moustafa2015unsw} in a lifelong learning scenario, showcasing adaptation to new attacks while retaining prior knowledge compared to static counterparts.
\end{itemize}



\section{Proposed Hierarchical D-SNN Methodology}

\begin{figure}[htbp]
    \centerline{\includegraphics[width=1.03\columnwidth]{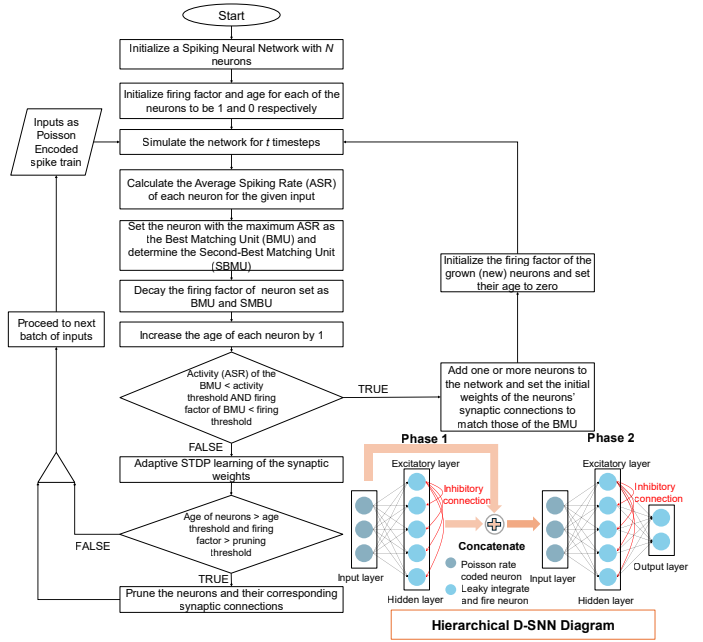}} 
    \caption{\footnotesize Flowchart illustrating the core D-SNN algorithm, including dynamic structural plasticity (growth/pruning) and Ad-STDP learning, as applied in the Phase 2 module. The inset shows the two-phase hierarchical SNN architecture (Phase 1 is static, Phase 2 is dynamic).}
    \label{fig:d-snn_flow_arch} 
    \vspace{-10pt}
\end{figure}

Our proposed approach utilizes a \textbf{Hierarchical D-SNN} architecture specifically designed for adaptive and efficient Network Intrusion Detection. The motivation for a hierarchical structure stems from several observations relevant to NIDS. Firstly, cyberattacks are often sparse events compared to the high volume of benign network traffic. A single, complex classifier processing all traffic is inefficient. Secondly, real-world network data is inherently imbalanced. A hierarchical design allows for efficient filtering and helps mitigate the challenges posed by this imbalance. Inspired by biological processing pathways \cite{ledoux2012rethinking} and the need for efficiency, our architecture decomposes the intrusion detection task into two cascaded SNN modules (illustrated in the inset of Fig.~\ref{fig:d-snn_flow_arch}): 

\noindent \textbf{Phase 1 (Attack Detection):} A lightweight SNN module acts as an initial filter. It employs a \textbf{static architecture with a fixed size of 100 neurons}. It processes incoming network features (encoded as Poisson spike trains \cite{diehl2015unsupervised}) to make a coarse determination: is the traffic potentially malicious or benign?

\noindent \textbf{Phase 2 (Attack Classification):} This module is activated only when Phase 1 flags potential malicious activity. It utilizes a \textbf{dynamic structure} capable of adaptation. It receives the original input features concatenated with the activity state (average spiking rate) of the Phase 1 excitatory neurons. Its task is to classify the specific type of attack detected.

Both modules are SNNs built with Leaky Integrate-and-Fire (LIF) neurons, leveraging their event-driven nature for potential energy savings \cite{sengupta2019going}. Lateral inhibition and homeostatic adaptive thresholds are used within each module's excitatory layer to promote neuron specialization and prevent dominance \cite{diehl2015unsupervised}.
Beyond the hierarchical structure, the core innovation lies in the network's dynamic nature, enabling adaptation and lifelong learning. Inspired by GWR principles and the need to combat catastrophic forgetting, the Phase 2 D-SNN dynamically adjusts its structure and synaptic plasticity. The complete workflow, including the hierarchical structure, dynamic adaptation, and learning, is depicted in Fig.~\ref{fig:d-snn_flow_arch}. 

\vspace{-10pt}

\subsection{Key Concepts for Dynamic Adaptation}

Several key metrics and concepts govern the dynamic behavior of the D-SNN module and adaptive learning -

\noindent \textbf{Average Spiking Rate (ASR):} A measure of a neuron's recent firing activity over a sliding time window, indicating its response to current input stimuli. 

\noindent \textbf{Best Matching Unit (BMU) \& Second-Best Matching Unit (SBMU):} For a given input, the BMU is the excitatory neuron with the highest ASR, representing the closest match in the network. The SBMU is the neuron with the second-highest ASR \cite{marsland2002self}.

\noindent \textbf{Firing Factor ($f_i$):} To balance plasticity and stability during lifelong learning, we introduce a neuron-specific firing factor. Intuitively, new neurons need to be highly adaptable to learn new patterns, while neurons that have already specialized in representing certain inputs should become more stable to retain that knowledge. This concept is inspired by the biological principle of \textit{habituation} (where neurons become less responsive to repeated stimuli) and GWR's habituation counter \cite{marsland2002self}. Our proposed firing factor ($f_i$) implements this idea by tracking a neuron's activity history and influencing its plasticity. It starts high (e.g., 1) for newly added neurons, promoting learning. As a neuron is frequently selected as the BMU or SBMU for inputs, indicating its successful integration and specialization, its firing factor decays over time according to $f_{i} = 1 - \frac{1}{\alpha_{i}} \left(1 - e^{-(\alpha_{i} \cdot n) / \tau_{ff}} \right)$, where  $n$ tracks the number of times the neuron was selected as BMU or SBMU, $\tau_{ff}$ is the decay time constant, and $\alpha_i$ is a neuron-specific rate constant. The SBMU's decay rate ($\alpha_{sbmu}$) is further scaled by its ASR relative to the BMU ($\alpha_{SBMU} = \alpha_{BMU} \cdot \frac{ASR_{SBMU}}{ASR_{BMU}}$). 

\noindent \textbf{Neuron Age:} A counter associated with each neuron, incremented over time (e.g., per mini-batch). It is used in conjunction with the firing factor for the pruning mechanism in Phase 2.

\subsection{Dynamic Structural Plasticity (Phase 2)}

The Phase 2 SNN module adapts its structure through neuron growth and pruning:

\noindent \textbf{Network Growth:} To accommodate new patterns without causing catastrophic forgetting, new excitatory neurons are added strategically. Growth is triggered when the BMU responds weakly (ASR $< a_{th}$, an activity threshold) to an input pattern that it \textit{should} recognize, indicated by its low, decayed firing factor ($f_{BMU} < f_{th}$, a firing threshold). A low firing factor signifies that the BMU has already specialized in learning previous patterns; forcing it to learn this new, poorly matched pattern could overwrite its existing knowledge. Therefore, the dual condition (low ASR and low $f_{BMU}$) identifies the need for a \textit{new} neuron to handle the novel pattern. This new neuron inherits weights similar to the BMU for rapid integration but starts with a high firing factor ($f_i=1$) and zero age, maximizing its initial plasticity specifically for learning the new pattern.

\noindent \textbf{Network Pruning:} To maintain efficiency and remove redundant units, neurons are pruned based on their age and activity history. If a neuron's age exceeds a maximum threshold ($age > age_{max}$) \textit{and} its firing factor remains consistently high ($f_i > p_{th}$, a pruning threshold), it suggests the neuron has failed to specialize or contribute meaningfully to pattern representation. Such neurons, along with their connections, are removed from the network.

\subsection{Learning Rules: STDP and Ad-STDP}

Synaptic weights are updated using different STDP-based rules in each phase. The static Phase 1 module employs standard STDP, where weight changes depend only on spike timing \cite{diehl2015unsupervised}. The dynamic Phase 2 module uses our novel Adaptive STDP (Ad-STDP) rule, a key contribution of this work. This rule introduces and incorporates the presynaptic neuron's firing factor ($f_i$) to modulate plasticity (Eq.~\ref{eq:adstdp}), balancing stability and adaptability.
\vspace{-10pt}
\begin{equation} \label{eq:adstdp}
\Delta w_{ij} =
\begin{cases}
A_+ \cdot f_i \cdot \exp(- \Delta t / \tau_{pre}) & \text{if } \Delta t > 0 \text{ (LTP)} \\
A_- \cdot f_i \cdot \exp(+ \Delta t / \tau_{post}) & \text{if } \Delta t < 0 \text{ (LTD)}
\end{cases}
\vspace{-5pt}
\end{equation}

Here, $\Delta t = t_{post} - t_{pre}$ is the relative timing difference between postsynaptic ($t_{post}$) and presynaptic ($t_{pre}$) spikes. $A_+$ and $A_-$ represent the maximum amplitudes for Long-Term Potentiation (LTP) and Long-Term Depression (LTD), respectively. $\tau_{pre}$ and $\tau_{post}$ are the time constants governing the STDP window for potentiation and depression. \textbf{Our work's novelty lies in modulating these updates with $f_i$:} High $f_i$ of new or inactive neurons allows larger weight updates, facilitating rapid learning. As a neuron becomes established ($f_i$ decays), the magnitude of weight updates decreases. This stabilizes learned representations and prevents new learning from drastically overwriting existing knowledge, effectively mitigating catastrophic forgetting. If $f_i$ drops very low (approaching a habituated state), plasticity is significantly reduced for that neuron's outgoing synapses.

\subsection{Semi-Supervised Labeling}

Following the unsupervised phase involving structural plasticity and \textbf{STDP/Ad-STDP learning}, a small amount of labeled data is used to assign functional labels to the excitatory neurons \cite{diehl2015unsupervised}. In Phase 1, neurons are labeled as `Attack' or `Benign' based on their maximal ASR response to corresponding labeled inputs. In Phase 2, neurons are assigned specific attack type labels (e.g., `DOS', `DDOS', etc.) using the same principle. This semi-supervised approach leverages the network's self-organization while minimizing the requirement for extensively labeled datasets, making it suitable for real-world scenarios where labeled data may be scarce.



\section{Experimental Setup and Results}

\subsection{Experimental Setup}

We evaluate our proposed \textbf{Hierarchical D-SNN} against a baseline Static Hierarchical SNN on the UNSW-NB15 NIDS dataset \cite{moustafa2015unsw}. Our version focuses on lifelong learning across six distinct attack classes selected from the original nine; the remaining three classes (e.g., Worms, Shellcode, Analysis) are excluded due to having too few samples to form meaningful sequential tasks in our lifelong learning protocol. The data is preprocessed using standard cleaning, scaling, and random forest feature selection (42 features), with an 8:1:1 train/validation/test split. To assess adaptation, we simulate a task-incremental \textbf{lifelong learning} scenario: the network is trained sequentially on distinct tasks, each introducing benign traffic and a new, disjoint set of attack types (e.g., Task 1: DOS/Scanning; Task 2: Backdoor/DDOS, etc.), without revisiting prior task data. This protocol mimics real-world adaptation needs without full retraining.
We use Python-based simulation framework with BindsNET \cite{hazan2018bindsnet}. Efficiency analysis use Intel Lava framework \cite{davies2021advancing} simulations. 

\subsection{Results}
\begin{figure}[htbp]
\centerline{\includegraphics[width=0.65\columnwidth]{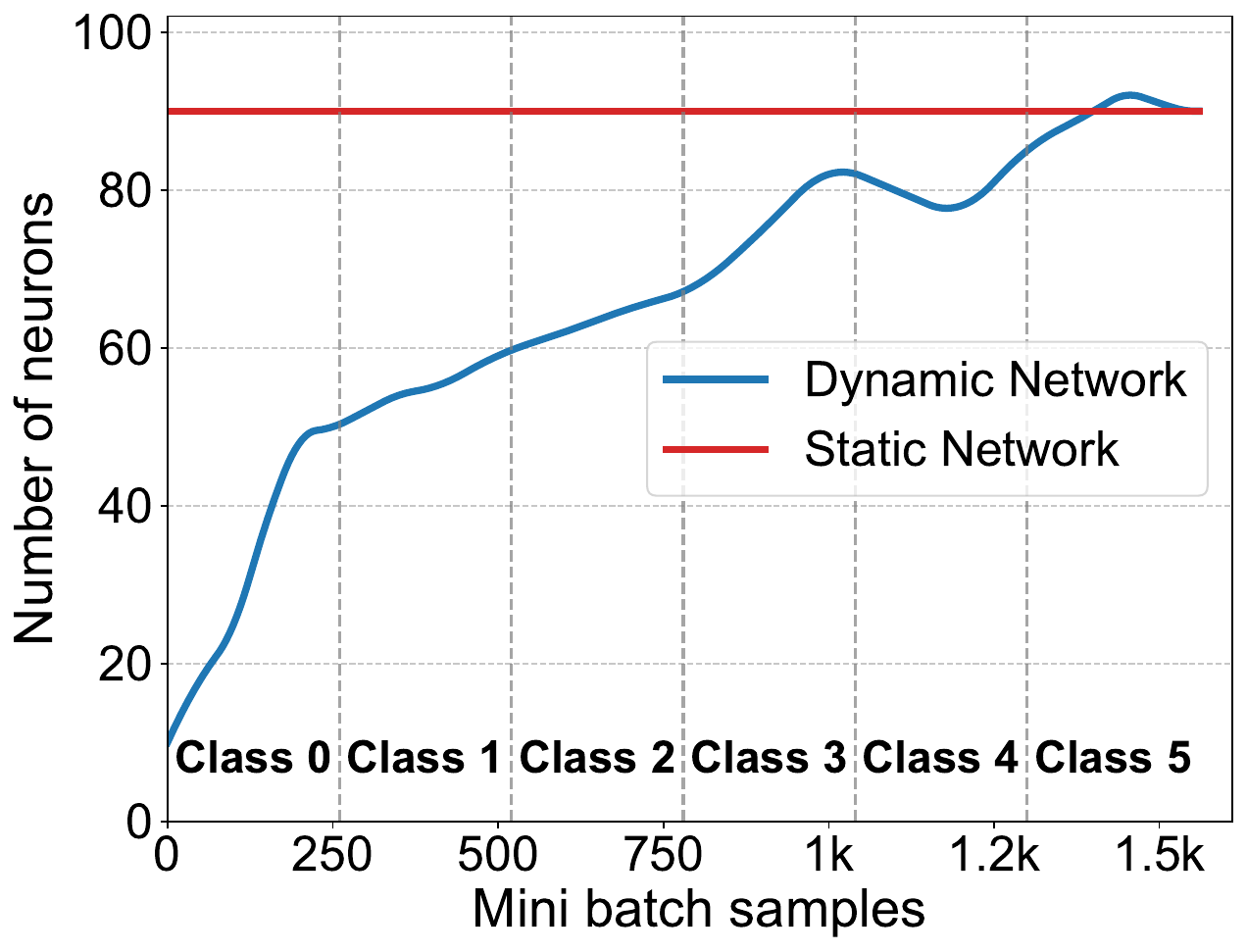}}
\vspace{-10pt} 
    \caption{\footnotesize Neuron count evolution in the D-SNN during lifelong learning.}
    \label{fig:neuron_growth}
    \vspace{-5pt}
\end{figure}

Compared to the static baseline which suffers significant performance degradation, the proposed Hierarchical D-SNN effectively adapts to new tasks and mitigates catastrophic forgetting, demonstrating the benefits of its dynamic structure and adaptive learning for lifelong operation.
The mechanism enabling this improved adaptation is visualized in Fig.~\ref{fig:neuron_growth}, tracking the structural evolution during the lifelong learning process. The network begins with few neurons and dynamically increases its size (growing to approx. $90$ neurons) as it encounters new information corresponding to different attack classes. This contrasts sharply with the static baseline's fixed capacity. 

\vspace{-10pt}
\begin{figure}[htbp]         \centerline{\includegraphics[width=0.85\columnwidth]{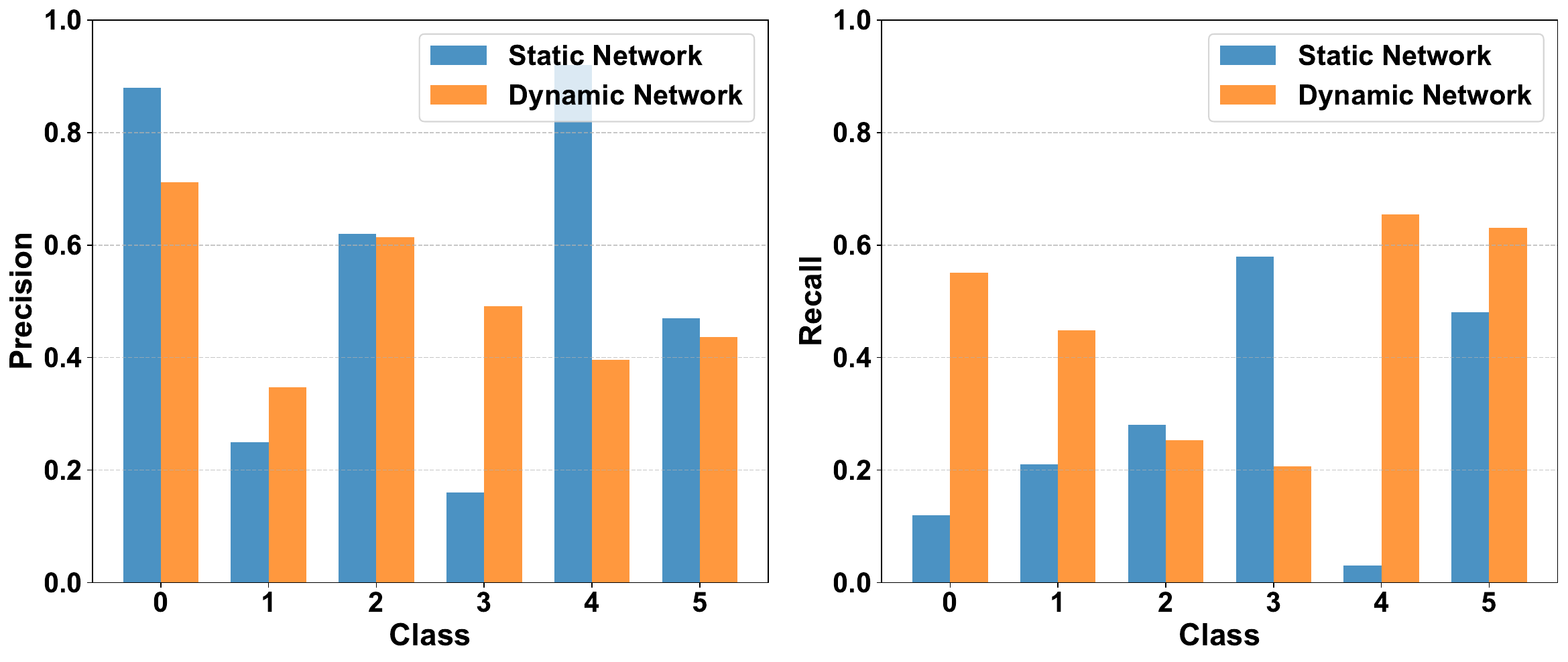}}
\vspace{-10pt} 
    \caption{\footnotesize Precision and Recall comparison per NIDS attack class for static vs. dynamic SNNs.}
    \label{fig:pr_comparison}
    \vspace{-5pt}
\end{figure}

The performance benefit of dynamic adaptation is reflected in Fig.~\ref{fig:pr_comparison}. Critically for NIDS, the dynamic network shows substantially higher recall for most attack classes (0, 1, 4, 5), indicating superior pattern learning and knowledge retention essential for lifelong learning. While precision varies and static recall is higher for Class 3, the overall recall trend supports the dynamic approach's adaptability. While recent static, supervised deep learning models report high multi-class classification accuracies on UNSW-NB15 (often exceeding $95$\% \cite{kassem2024machine}), our \textbf{Hierarchical D-SNN addresses the distinct challenges of lifelong learning using a semi-supervised SNN approach and is the first to report performance of a neuromorphic algorithm in this domain.} 
Based on the Phase 1 detection accuracy ($94.3$\%) and Phase 2 classification accuracy ($66.3$\%), weighted by the proportion of benign ($72.5$\%) and attack ($28.5$\%) traffic, the estimated overall system accuracy is approximately $85.3$\%. This significantly outperforms the static SNN baseline, whose overall accuracy under the same conditions is estimated at 80.0\% (using a Phase 2 static accuracy of 46.6\%). 

\vspace{-10pt}

Efficiency analysis using the Intel Lava framework highlights the benefits of our approach. The Hierarchical D-SNN operates with high sparsity, indicated by very low average inference spike rates per neuron ($\sim0.0008$ for the Phase 1 filter, $\sim0.001-0.002$ for the dynamic classifier over 200 timesteps). This inherent sparsity, particularly in the initial detection stage, is significantly greater than typical ANN-SNN conversion techniques \cite{sengupta2019going} and directly contributes to potential energy savings, as computation is primarily event-driven. 


\section{Conclusions}
Our Hierarchical D-SNN integrates structural plasticity with adaptive Ad-STDP learning, enabling lifelong NIDS capabilities that mitigate catastrophic forgetting and improve pattern learning over static SNNs. Its advantages include hierarchical efficiency, inherent sparsity suitable for neuromorphic hardware, and reduced label dependency through semi-supervised learning. 
As the first hierarchical D-SNN combining these techniques for semi-supervised continual learning in NIDS, this architecture offers a promising direction for robust, energy-efficient cybersecurity, despite overheads associated with dynamic network growth/pruning. Future work will focus on on-chip learning, Ad-STDP refinement, and evaluation on more complex datasets.



\bibliographystyle{IEEEtran}

\end{document}